\documentclass[11pt]{article}

\usepackage[T1]{fontenc}
\usepackage[utf8]{inputenc}
\usepackage{lmodern}
\usepackage{microtype}

\usepackage[margin=1in]{geometry}

\usepackage{amsmath,amssymb}

\usepackage{graphicx}
\usepackage{pdflscape}   
\usepackage{float}       
\usepackage{makecell}    
\usepackage{algorithm}
\usepackage{algpseudocode}

\usepackage{longtable,booktabs,array}
\usepackage{calc}
\usepackage{etoolbox}
\makeatletter
\patchcmd\longtable{\par}{\if@noskipsec\mbox{}\fi\par}{}{}
\makeatother
\IfFileExists{footnotehyper.sty}{\usepackage{footnotehyper}}{\usepackage{footnote}}
\makesavenoteenv{longtable}

\usepackage{authblk}

\usepackage{xurl}
\usepackage{hyperref}
\hypersetup{
  colorlinks=true,
  linkcolor=blue,
  citecolor=blue,
  urlcolor=blue,
  pdftitle={NVAITC AI Scientist: A Governed End-to-End Research System --- A Hypertension GWAS Case Study}
}

\usepackage[numbers,sort&compress]{natbib}

\title{NVAITC AI Scientist: A Governed End-to-End Research System\\---
A Hypertension GWAS Case Study}

\author[1,*]{Eddie Huang}
\author[1]{Ken Liao}
\author[1]{Iven Fu}
\author[1]{Yang-Hsien Lin}
\author[1]{Chao-Shun Zhan}
\author[1]{Andy Liao}
\author[1]{Virginia Chen}
\author[1]{Johnson Sun}
\author[1]{Pika Wang}
\author[1]{Richard Huang}
\author[1]{Jiun-Cheng Jiang}
\author[2,3]{Ting-Yuan Liu}
\author[2,3,4]{Hsing-Fang Lu}
\author[5]{Ray Y. Lee}
\author[2]{Chi-Chou Liao}
\author[1]{Simon See}
\author[2,6,7,8,**]{Fuu-Jen Tsai}

\affil[1]{NVIDIA AI Technology Center (NVAITC), NVIDIA Corporation}
\affil[2]{Department of Medical Research, China Medical University Hospital, Taichung 40402, Taiwan}
\affil[3]{Master Program for Digital Health Innovation, China Medical University, Taichung 406040, Taiwan}
\affil[4]{Laboratory for Statistical and Translational Genetics, RIKEN Center for Integrative Medical Sciences, Yokohama, Japan}
\affil[5]{AI-Driven Genomic Medicine and Drug Discovery Lab, China Medical University Hospital, Taichung 40402, Taiwan}
\affil[6]{School of Chinese Medicine, China Medical University, Taichung 40402, Taiwan}
\affil[7]{Division of Pediatric Genetics, Children's Hospital of China Medical University, Taichung 40447, Taiwan}
\affil[8]{Department of Medical Laboratory Science and Biotechnology, Asia University, Taichung 41354, Taiwan}
\affil[*]{Corresponding author: \href{mailto:tzungchih@nvidia.com}{tzungchih@nvidia.com}}
\affil[**]{Corresponding author: \href{mailto:000704@tool.caaumed.org.tw}{000704@tool.caaumed.org.tw}}

\date{}

\begin{document}
\maketitle

\begin{abstract}
Agentic research systems are emerging as a new paradigm for coordinating
scientific workflows beyond isolated model inference, code generation,
or statistical analysis. However, deployment in institutional biomedical
environments requires governed mechanisms for research planning, data
access, workflow orchestration, evidence tracking, reproducibility, and
human oversight. We present NVAITC AI Scientist (NAIS), a governed
end-to-end agentic research system designed to support domain-general
scientific workflows while keeping protected data within institutional
privacy boundaries. NAIS integrates proposal review, execution planning,
governed computational routing, reproducible workflow orchestration,
evidence generation, and scientist-in-the-loop oversight. We validate
NAIS in a real-world hypertension genome-wide association study (GWAS)
using hospital-linked genotype and electronic health record (EHR) data from
286,422 individuals under an aggregate-only data policy. The agent
planned cohort extraction, orchestrated GWAS execution, generated
quality-control summaries and visualizations, and drafted
publication-oriented outputs. Systematic comparison with independently
curated expert analyses showed that human-AI review identified phenotype
discrepancies and enabled iterative refinement of the hypertension
definition. After team-directed reconciliation, the agent-orchestrated
GWAS reproduced established hypertension-associated loci, including
FGF5, ATP2B1, CNNM2, FTO, and GRB14, with the strongest signal at FGF5
reaching $-\log_{10} p \approx 70$. As a secondary demonstration beyond GWAS, NAIS
also supported a drug-induced liver injury prediction workflow,
achieving a multimodal graph neural network area under the curve (AUC) of 0.842. These results
demonstrate that governed agentic research systems can support scalable
AI-assisted biomedical discovery while producing scientifically reliable
outputs comparable to expert-led workflows.

\end{abstract}

\section{Introduction}

Agentic research systems are beginning to extend artificial intelligence
from isolated tasks, such as model inference, code generation,
literature summarization, or statistical analysis, toward the
coordination of complete scientific workflows \cite{gottweis2026,lu2024,weng2024,mitchener2025,gao2026,liu2026auto,liu2024autogl,companionai2026,zhipuai2026}. These systems
promise to assist researchers across planning, execution,
interpretation, and manuscript preparation. However, most demonstrations
of autonomous or semi-autonomous research agents have focused on open
data, simulated environments, local software execution, or benchmark
tasks. Translating agentic research into institutional biomedical
settings introduces a different class of requirements: protected data
must remain within approved boundaries, computational actions must be
auditable, workflows must be reproducible, and scientific decisions must
remain under human oversight.

Biomedical research linked to electronic health record (EHR) data, genotype
banks, and hospital data warehouses highlights these challenges. Even
when mature statistical tools are available, real-world studies depend
on difficult upstream decisions, including cohort definition, phenotype
harmonization, data-quality assessment, and sensitivity analysis
\cite{tam2019,carroll2012}. In genome-wide association studies (GWAS), for example,
blood-pressure measurements, diagnosis codes, and medication records may
produce different hypertension labels, and phenotype choices can
substantially affect association results. These issues cannot be solved
by a language model that merely emits Structured Query Language (SQL) or analysis code. They require
a governed research environment in which agents can plan and orchestrate
analyses while data access, execution, artifact release, and scientific
interpretation remain controlled.

\begin{figure}[htbp]
\centering
\includegraphics[width=\linewidth]{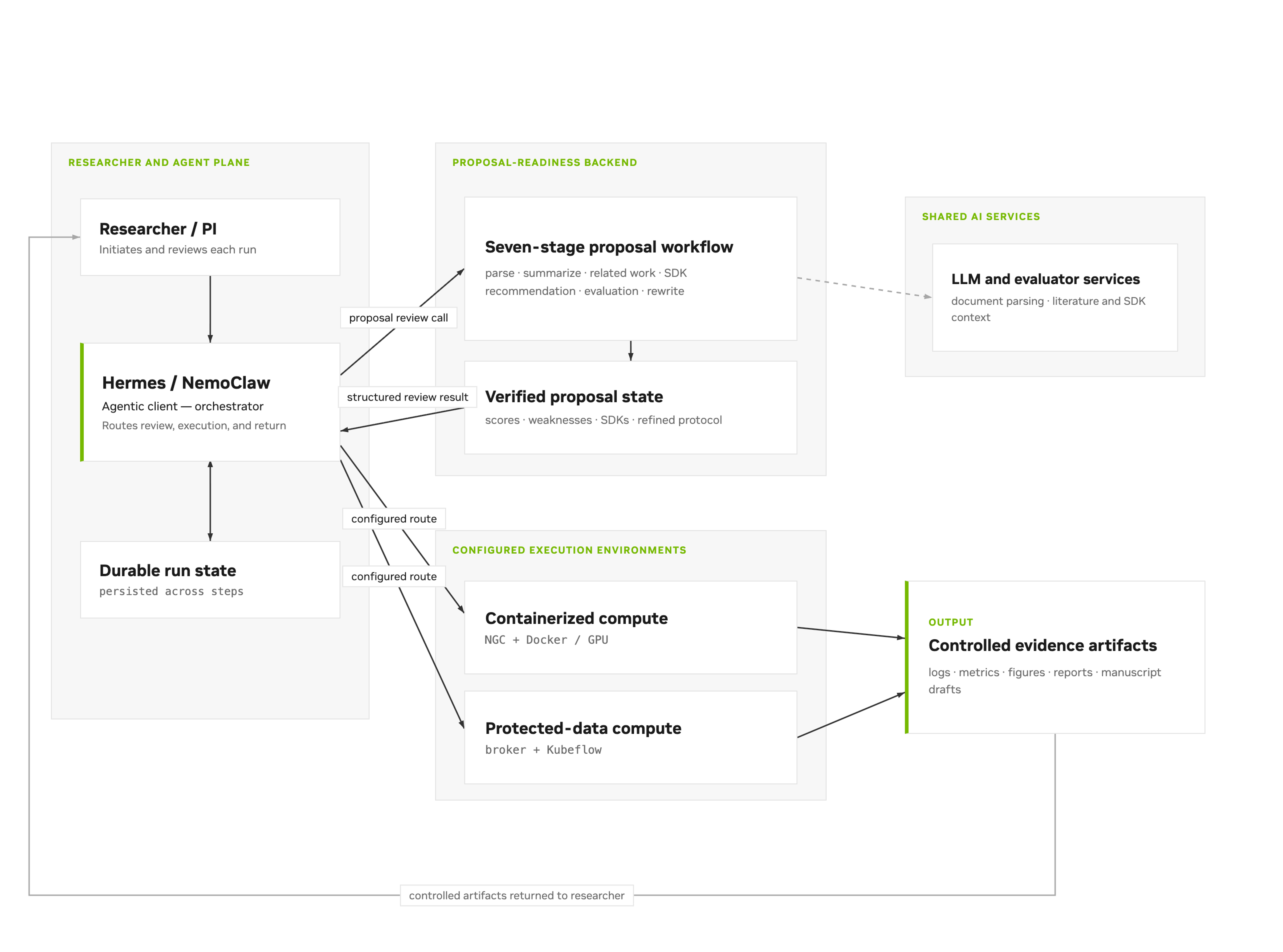}
\caption{NAIS end-to-end architecture. Optional proposal review feeds
structured state to NemoClaw, which executes via the NVIDIA GPU Cloud container path
or the broker/Kubeflow governed path, producing auditable evidence and
manuscript artifacts.}
\label{fig:architecture}
\end{figure}

We present NVAITC AI Scientist (NAIS), a governed end-to-end agentic
research system designed to support scientific workflows in
institutionally constrained environments. NAIS combines durable research
state, proposal-level planning, agentic execution, brokered data access,
reproducible workflow orchestration, evidence tracking, and
scientist-in-the-loop oversight. Rather than granting an agent
unrestricted access to protected data, NAIS separates reasoning and
orchestration from data governance: the agent submits approved actions
through controlled interfaces, protected data remain inside
institutional infrastructure, and only governed artifacts such as
aggregate summaries, quality-control metrics, plots, logs, and
manuscript-oriented evidence are returned. Figure~\ref{fig:architecture} illustrates the overall architecture, showing how proposal review, agentic execution, and governed data access compose into an end-to-end research system.

The primary validation of NAIS is a real-world hypertension GWAS
conducted on hospital-linked genotype and EHR data from the China
Medical University Hospital (CMUH) HiGenome Genomic Bank. In this
deployment, NemoClaw, the agentic execution component of NAIS, planned
cohort extraction, submitted governed analysis specifications,
orchestrated PLINK2 GWAS workflows through brokered Kubeflow execution,
retrieved aggregate analysis artifacts, and drafted manuscript sections.
The study involved 286,422 individuals and operated under an
aggregate-only data policy, allowing the agent to assist research near
protected clinical data without direct access to participant-level
records.

This case study also illustrates why governed agentic research requires
human-AI collaboration rather than unconstrained autonomy. During cohort
construction, comparison between agent-derived labels and independently
curated expert phenotypes revealed systematic discordance in
hypertension classification. Human review, medication audit, and
team-directed phenotype reconciliation refined the cohort definition
before final analysis. After reconciliation, the agent-orchestrated GWAS
reproduced established hypertension-associated loci, including FGF5,
ATP2B1, CNNM2, FTO, and GRB14, matching independently curated expert
results in locus identification and signal direction. These findings
support the central claim of this work: governed agentic systems can
execute end-to-end biomedical research workflows while preserving
institutional privacy boundaries and producing results comparable to
expert-led workflows.

This paper makes four contributions. First, we built NAIS, a governed
end-to-end architecture for agentic research in protected biomedical
environments, including durable state management, brokered execution,
auditability, and aggregate-only artifact return. Second, we validate
the governed execution path through a 286,422-participant hypertension
GWAS with independent expert comparison, phenotype-discordance analysis,
medication audit, and final replication of established loci. Third, we
characterize the role of scientist-in-the-loop oversight in converting
agent-generated workflows into research-grade analyses, showing that
human-AI refinement was essential for phenotype reconciliation and
scientific validation. Fourth, we provide supporting demonstrations of
broader NAIS capabilities, including proposal review, containerized
compute workflows, and a secondary drug-induced liver injury prediction
case study. Together, these results position NAIS as a blueprint for
scalable AI-assisted biomedical research under institutional governance.

\section{Related Work}

Agentic research systems, automated peer review, and workflow orchestration frameworks each inform the design of NAIS. We review representative work across these areas and position NAIS within this landscape.

\subsection{Agentic Systems for Scientific Discovery}

Recent agentic research systems have moved beyond single-step assistance
toward multi-stage scientific workflows that include hypothesis
generation, literature grounding, experimental planning, code execution,
result interpretation, and manuscript drafting. Co-Scientist uses a
multi-agent generate--debate--evolve process for biomedical hypothesis
generation and has been validated through experimental follow-up in
areas including drug repurposing and liver fibrosis \cite{gottweis2026}. The AI
Scientist and CycleResearcher demonstrate closed-loop automation of
machine-learning research, including idea generation, experiment
execution, paper writing, and automated review \cite{lu2024,weng2024}. Kosmos extends
long-horizon data-driven discovery through a structured world model that
coordinates literature search and data analysis over many agent rollouts
\cite{mitchener2025}. AutoScientists and AutoResearchClaw further emphasize
decentralized agent teams, dead-end recovery, self-healing execution,
cross-run learning, and human-AI collaboration across benchmarked
scientific tasks \cite{gao2026,liu2026auto}.

Together, these systems establish that language-model agents can
coordinate substantial portions of the scientific workflow. However,
most existing demonstrations emphasize open or public datasets,
benchmark tasks, local software execution, simulated review, or
hypothesis generation followed by separate validation. They generally do
not address the institutional constraints that arise when agents operate
near protected clinical data, where raw records cannot be exposed to
unconstrained reasoning loops and where data access, execution, artifact
release, auditability, and human approval must be governed. NAIS is
positioned in this complementary space: not primarily as an autonomous
hypothesis generator, but as a governed agentic research system for
orchestrating real biomedical workflows within institutional data
boundaries.

\subsection{Clinical and Genomic Workflow Orchestration}

Kubernetes and Kubeflow Pipelines provide foundations for containerized,
reproducible ML workflows \cite{kubeflow2026,kubernetes2026}. Hospital GWAS additionally
requires narrow action interfaces, cohort SQL against complex data warehouse (DWH)
schemas, PLINK-compatible phenotype files, and phenotype-definition
sensitivity analysis. NAIS binds these through a broker v2 run application programming interface (API),
in-boundary SQL cohort-extraction and GWAS analysis pipelines, and a
GWAS orchestration skill that coordinates analysis-specification
submission, human approval gates, job polling, and artifact retrieval
under an aggregate-only policy.

\subsection{LLM Verification and Proposal Review}

Forward-only large language model (LLM) pipelines can silently drop details or return weakly
grounded assessments. Liang et al.\ \cite{liang2023}, Liu and Shah \cite{liu2020}, and
related work study LLM peer review as a verification strategy.
NAIS applies this pattern by wrapping each generative stage with evaluator
LLMs, catching 852 deficiencies across 18 production proposal-review
runs with varying fields such as biological, industrial, quantum, and
robotics sciences (Appendix A, Tables A1--A3). Thorne et al.\ \cite{thorne2026}
evaluated LLM grant review on EPSRC proposals. These patterns complement
but do not replace governed execution validation on real hospital data.

\subsection{Comparative Positioning}

Table \ref{tab:comparison} situates NAIS within the emerging landscape of agentic research
systems. While several peers emphasize hypothesis generation and
multi-round evolution, NAIS occupies a complementary governed-execution
niche defined by brokered API access to protected clinical and genotype
data and preserved audit trails. The hypertension GWAS case study provides primary
end-to-end validation absent from systems that either lack governance
disclosures or validate primarily on synthetic or public-data
benchmarks. Rather than generating novel hypotheses autonomously, NAIS
demonstrates its strength in orchestrating domain pipelines near
protected data with scientist-in-the-loop reconciliation.

\begin{landscape}
{\footnotesize
\renewcommand{\arraystretch}{2.0}
\begin{longtable}[]{@{}
  >{\raggedright\arraybackslash}p{0.08\textheight}
  >{\raggedright\arraybackslash}p{0.11\textheight}
  >{\raggedright\arraybackslash}p{0.11\textheight}
  >{\raggedright\arraybackslash}p{0.09\textheight}
  >{\raggedright\arraybackslash}p{0.12\textheight}
  >{\raggedright\arraybackslash}p{0.10\textheight}
  >{\raggedright\arraybackslash}p{0.12\textheight}
  >{\raggedright\arraybackslash}p{0.09\textheight}
  >{\raggedright\arraybackslash}p{0.09\textheight}@{}}
\caption{Capability comparison across agentic research systems (bold text indicates leads in each dimension).}\label{tab:comparison}\tabularnewline
\toprule
Dimension &
NAIS &
Co-Scientist \cite{gottweis2026} &
Kosmos \cite{mitchener2025} &
AutoScientists \cite{gao2026} &
AI Scientist \cite{lu2024} &
\makecell{Auto\\ResearchClaw} \cite{liu2026auto} &
Feynman \cite{companionai2026} &
\makecell{Auto\\Claw} \cite{zhipuai2026} \\
\midrule
\endfirsthead
\toprule
Dimension &
NAIS &
Co-Scientist \cite{gottweis2026} &
Kosmos \cite{mitchener2025} &
AutoScientists \cite{gao2026} &
AI Scientist \cite{lu2024} &
\makecell{Auto\\ResearchClaw} \cite{liu2026auto} &
Feynman \cite{companionai2026} &
\makecell{Auto\\Claw} \cite{zhipuai2026} \\
\midrule
\endhead
\bottomrule
\endlastfoot
Iterative experiment loop &
Single-pass execution &
\textbf{Tournament evolution} &
\textbf{20-cycle + world model} &
\textbf{Multi-round + reorganization} &
Closed-loop ideation $\to$ experiment $\to$ paper &
Pivot/Refine + cross-run evolution &
Bounded replication loop &
Multi-step, no formal loop \\

Literature grounding &
Related-work in proposal phase &
\textbf{Evidence + scientist input} &
\textbf{$\sim$1,500 papers/run} &
Shared record only &
Semantic Scholar in pipeline &
Web search + debate context &
\textbf{Citation verification} &
Web search summaries \\

Security \& governance &
\textbf{Zero-trust, protected health information isolation, audit trail} &
Not disclosed &
Not disclosed &
None &
None &
None &
None &
On-device only \\

Stagnation recovery &
Evaluator retry &
\textbf{Tournament elimination} &
Parallel rollouts &
\textbf{Dead-end detection} &
Within-run code revision &
\textbf{Self-healing + Pivot/Refine} &
Iterative refinement &
Retries until done \\

Hypothesis generation &
Evaluates (does not generate) &
\textbf{Generate-debate-evolve} &
\textbf{Updated each cycle} &
Implicit (team proposals) &
\textbf{Autonomous idea generation} &
\textbf{Multi-agent debate} &
Plans from literature &
Not a focus \\

Real-world validation &
\textbf{GWAS + drug-induced liver injury + expert validation} &
\textbf{Wet-lab (AML, liver fibrosis)} &
\textbf{7 discoveries, 3 reproduced} &
24 biomedical benchmarks (+8.3 pp) &
Automated ML papers &
ARC-Bench (+54.7\% vs AI Scientist v2) &
Paper replication &
Browser/GUI tasks \\
\end{longtable}
}
\end{landscape}

\section{The NVAITC AI Scientist System}

NAIS is designed to support the full research lifecycle on protected
clinical and genotype data, from proposal review and literature
grounding through governed pipeline execution to auditable evidence and
manuscript artifacts. The system is built around three design
principles: (1) durable state rather than ephemeral chat, allowing
workflows to survive failures and multi-day job queues, (2) brokered
data access rather than direct protected health information (PHI) exposure, allowing agents to work
near sensitive data without becoming an exfiltration channel, and (3)
scientist-in-the-loop oversight at each stage, ensuring that AI augments
rather than replaces researcher judgment. This section describes each
component, while how these capabilities played out in practice are
documented in subsequent sections.

\subsection{End-to-End Architecture}

NAIS organizes three cooperating components: (1) an optional
proposal-readiness service implementing a seven-stage NeMo Agent Toolkit
workflow, (2) NemoClaw as the agentic client maintaining durable run
state, and (3) an execution layer binding verified protocol to container
discovery or brokered Kubeflow workflows, producing auditable evidence
artifacts (\emph{Figure~\ref{fig:architecture}}). The two execution routes differ in
environment assumptions, agent actions, and returned artifacts
(\emph{Table~B1}, Appendix B).

\begin{figure}[htbp]
\centering
\includegraphics[width=\linewidth]{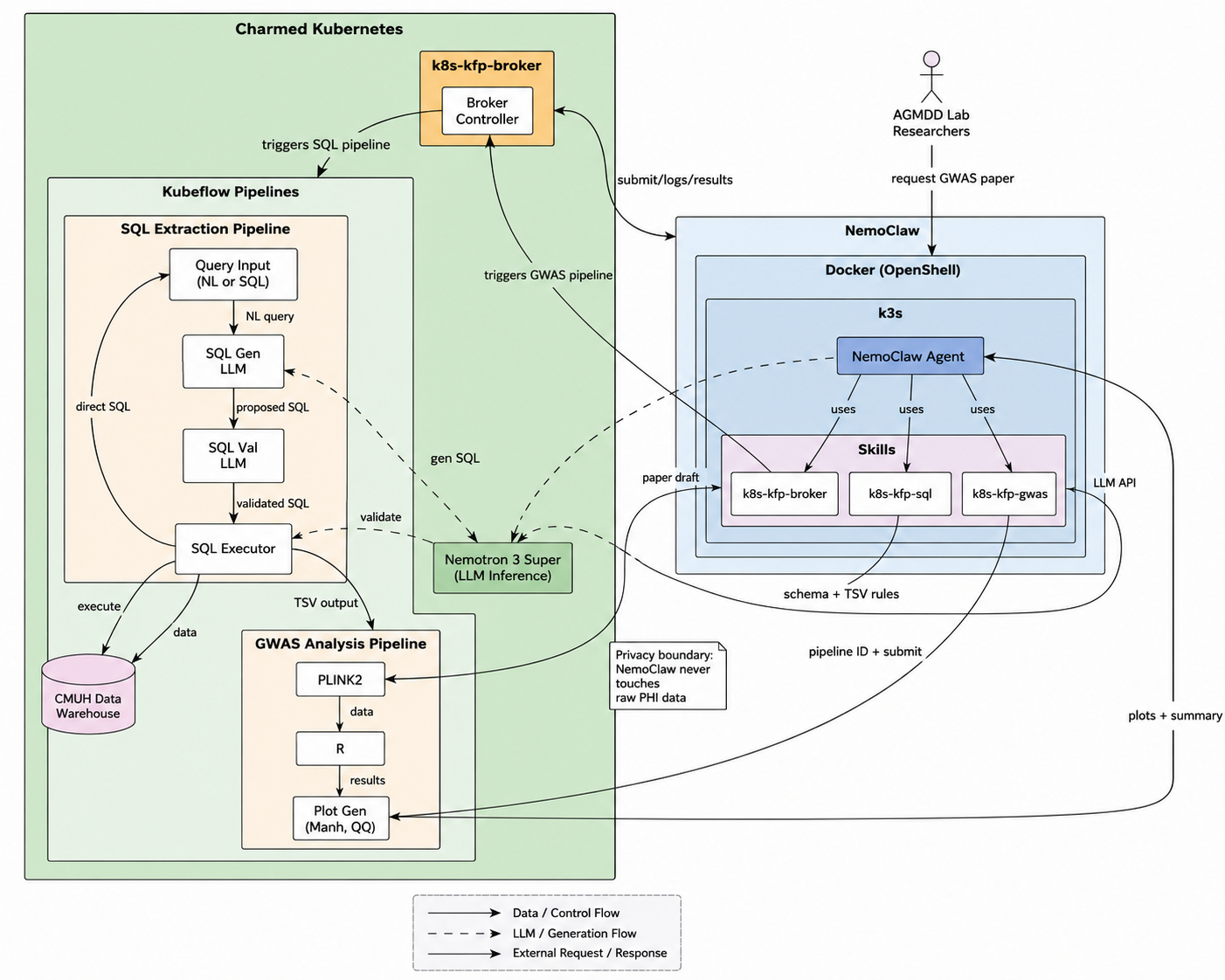}
\caption{Governed execution path validated in our hypertension GWAS
deployment. A team member issues a natural-language request via
Telegram. NemoClaw runs in an OpenShell sandbox and submits a structured
analysis specification to Broker. Broker launches SQL cohort extraction
and PLINK2 GWAS analysis in isolated Kubeflow containers. Only aggregate
summaries, quality control (QC) metrics, plots, and manifests return to the agent. Raw PHI
remains in the authorized data warehouse (DWH) vault, with only Broker permitted to access it. External
network access is blocked by default (HTTP 403) until a team member
approves a specific connection.}
\label{fig:governed}
\end{figure}

The execution layer treats reviewed proposals as durable state rather
than chat history: container selection, generated code, broker run IDs,
logs, metrics, repair notes, and manuscript sections persist across
failures and multi-day workflows. When jobs fail, the agent inspects
broker events, pod logs, and pipeline status before resubmitting,
behavior we observed repeatedly during deployment (broker connectivity
issues, duplicate sample IDs, empty generalized linear model (GLM) outputs, genotype path
debugging). \emph{Figure~\ref{fig:governed}} details the governed path we validated in
the hypertension GWAS deployment.

\subsection{Governed Execution Path}

In the governed variant, NemoClaw is limited to approved HTTP actions,
principally broker-mediated requests for GWAS and polygenic risk score runs, SQL cohort
extraction, schema discovery, cohort materialization, job launch and
polling, and artifact retrieval. It does not receive direct network file system, persistent volume claim, or
PHI access. The agent follows a standard orchestrated workflow: draft an
analysis specification, obtain team approval on the SQL cohort plan when
needed, materialize the cohort through the broker, launch GWAS
execution, monitor job status, and retrieve Manhattan/quantile-quantile (QQ) plots and GLM
outputs through controlled manifest URLs.

Under the default data-exposure policy, the agent receives structure,
row counts, split sizes, null counts, numeric summaries, QC summaries,
execution summaries, failure-stage diagnostics, plots, and file lists,
but not raw SQL result rows, linkable patient identifiers, or free-text
clinical notes. This policy aligns with our requirement that AI assist
research without becoming a PHI exfiltration channel.

NemoClaw runs on Nemotron-3 Super (120B parameters) with 32,768-token
context, deployed on-premise. Its capabilities are composable rather
than fixed: broker-mediated pipeline submission, governed SQL cohort
extraction, GWAS execution, literature search, document parsing, and
model development can be combined as needed for each study, more like a
scientific toolbox than a single-purpose chatbot.

\subsection{Containerized Compute Path}

For accessible compute where data can leave institutional boundaries,
NemoClaw discovers and launches curated graphics processing unit (GPU) containers, generates
runnable code, and records provenance. Focused reruns across tabular ML,
deep learning inference optimization, and medical imaging segmentation
demonstrate evidence-traceable execution with container provenance,
logs, metrics, and reproducibility records (\emph{Table~C1}, Appendix C).
This path complements the broker/Kubeflow route.

\subsection{Proposal Review}

The proposal-readiness pipeline accepts PDF or text proposals, parses
them with a multimodal document-ingestion library, summarizes, searches
Semantic Scholar/arXiv, recommends NVIDIA SDKs, evaluates
novelty/technical merit/business value/ethics, and rewrites into the
NVAITC template. A reusable self-evaluating agent wrapper checks each
generative stage with evaluator LLMs (Algorithm 1, Appendix A). Across
18 production runs, evaluators detected 852 output deficiencies before
delivery (Table A2). This upstream stage proves NAIS end-to-end
capability.

\section{Agent Development}

When we first started using NemoClaw, the agent could converse about
genomics but could not yet reliably execute a hospital GWAS. The
hypertension case study in section 5 was the culmination of several
weeks of iterative work in which capability grew through real projects,
not a single turnkey deployment.

We began with a quantitative HbA1c GWAS on a hospital data warehouse.
NemoClaw proposed SQL extraction and a PLINK2 linear model, but early
attempts surfaced basic gaps: incorrect command sequencing, wrong
genotype paths, and Manhattan plots that did not reflect valid
association output. At the same time, the governed environment blocked
outbound network access by default, so the agent could not retrieve
PubMed or NCBI resources until we approved specific connections. Those
early failures clarified what the platform still needed: durable run
state, broker-mediated pipeline submission, and explicit human approval
before each cohort materialization.

Over the following weeks, successive projects expanded what the agent
could do. For a type 2 diabetes ethnicity review, NemoClaw searched the
literature, applied a PRISMA workflow, and produced manuscript text and
figures, demonstrating that scientific communication tasks were within
reach once external access was granted. For BMI and lipid GWAS runs, the
agent learned to navigate the hospital warehouse tables covering lab
results, diagnoses, and medications, diagnose duplicate-sample failures,
and resubmit corrected phenotype files through the broker's SQL
extraction pipeline. Each project left behind reusable skills,
orchestration patterns, and broker run configurations that carried
forward to the next study.

By the time we launched the hypertension GWAS, NemoClaw could draft an
analysis specification, obtain our approval on the SQL plan, materialize
a cohort, submit Kubeflow GWAS jobs on the Taiwan Precision Medicine Initiative (TPMI) imputed 50W library,
retrieve Manhattan and QQ plots from the broker manifest, and draft
methods and results sections. What remained difficult was not pipeline
execution but phenotype design. The agent initially classified
hypertension from laboratory blood-pressure values, while our clinical
genomics colleagues defined cases from ICD-10 codes and antihypertensive
prescriptions. Resolving that disagreement, through discordance
analysis, medication audit, and a directed rerun, became the central
validation result.

This progression illustrates a broader point about agentic research on
protected data: the model's reasoning improved less from prompt tuning
alone than from repeated engagement with real warehouse schemas, failing
pipelines, and team correction. Persistent broker state and composable
skills mattered as much as context length.

\section{Case Study: Hypertension GWAS Deployment}

The deployment provides the main validation evidence for NAIS. Over
several months, NemoClaw orchestrated multiple genomics and
drug-discovery tasks using a hospital genomic biobank. These efforts
began with early GWAS runs that exposed capability gaps and culminated
in a hypertension GWAS involving 286,422 individuals, with results
comparable to independently curated expert analyses. This section
focuses on the hypertension GWAS, covering cohort construction,
phenotype discordance, iterative human-AI reconciliation, and final
results.

The GWAS case study used de-identified genotype and phenotype data. The
study protocol was reviewed and approved by the Research Ethics
Committee of China Medical University Hospital (approval no.\
CMUH111-REC1-176). All analyses were performed within the institution's
protected computing environment, and no participant-level data were
transferred outside the approved environment.

\subsection{Study Setting and Research Goal}

Our team interacts with NemoClaw through a Telegram workspace, issuing
natural-language research goals that the agent translates into broker
analysis specifications, SQL cohort plans, and pipeline submissions. We
retain approval authority over SQL plans and analysis launches, while
the agent executes approved workflows and returns aggregate artifacts.

The hypertension GWAS goal was to estimate genome-wide associations for
hypertension in the hospital biobank, produce publication-quality
Manhattan and QQ plots, and draft methods/results text, all while
operating entirely within governed boundaries.

\subsection{Cohort Construction and Agent Workflow}

This study was approved by the institutional review board of the partner
hospital (approval no.\ CMUH111-REC1-176). All participants provided
written informed consent.

We defined hypertension cases using ICD-10 codes I10--I15 and/or
antihypertensive medication prescriptions, yielding 109,782 cases and
176,640 controls among 286,422 genotyped individuals (38.3\% prevalence).
This expert reference definition is distinct from an early agent
phenotype that relied on laboratory blood-pressure thresholds, which
produced discordant labels for 3,950 subjects. The final agent GWAS
adopted the ICD-10/medication definition after team-directed
reconciliation.

NemoClaw drafted an analysis specification, submitted it to broker v2,
and orchestrated: (1) SQL cohort extraction via the broker's SQL
extraction pipeline, (2) phenotype tab-separated values (TSV) generation with fields including
individual ID, hypertension label, systolic blood pressure (SBP) and diastolic blood
pressure (DBP), (3) GWAS submission via the broker's GWAS pipeline with
PLINK2 logistic regression, (4) per-chromosome GLM outputs merged into a
combined genome-wide association-results file, (5) Manhattan and QQ plot
generation.

Genotypes were derived from the 500k imputed array using the Haplotype
Reference Consortium panel. Post-imputation quality control
excluded variants with imputation INFO score below 0.8, minor allele
frequency below 0.01, or Hardy-Weinberg equilibrium $p$-value below
$1\times10^{-6}$. Genome-wide association testing used PLINK2 v2 logistic
regression with sex, age at test, and the first ten principal components
(PC1--PC10) as covariates to control for population stratification.
Phenotype labels were coded 1 (control) and 2 (case) rather than 0/1
because PLINK2 interprets 0 as a missing value. An earlier 0/1 encoding
had silently excluded all samples from the analysis until the agent
identified and corrected the artifact.

\begin{table}[htbp]
\centering
\caption{Cohort characteristics (expert reference phenotype, post-QC).}
\label{tab:cohort}
\begin{tabular}{ll}
\toprule
Characteristic & Value \\
\midrule
Total individuals & 286,422 \\
Hypertension cases & 109,782 (38.3\%) \\
Normotensive controls & 176,640 (61.7\%) \\
Mean age (years) & $58.2 \pm 14.7$ \\
Cases: mean SBP / DBP (mmHg) & $180.2 \pm 24.9$ / $107.1 \pm 15.7$ \\
Controls: mean SBP / DBP (mmHg) & $117.5 \pm 4.3$ / $78.2 \pm 3.3$ \\
\bottomrule
\end{tabular}
\end{table}

\subsection{Discordance Analysis and Iterative Refinement}

\begin{figure}[htbp]
\centering
\includegraphics[width=0.8\linewidth]{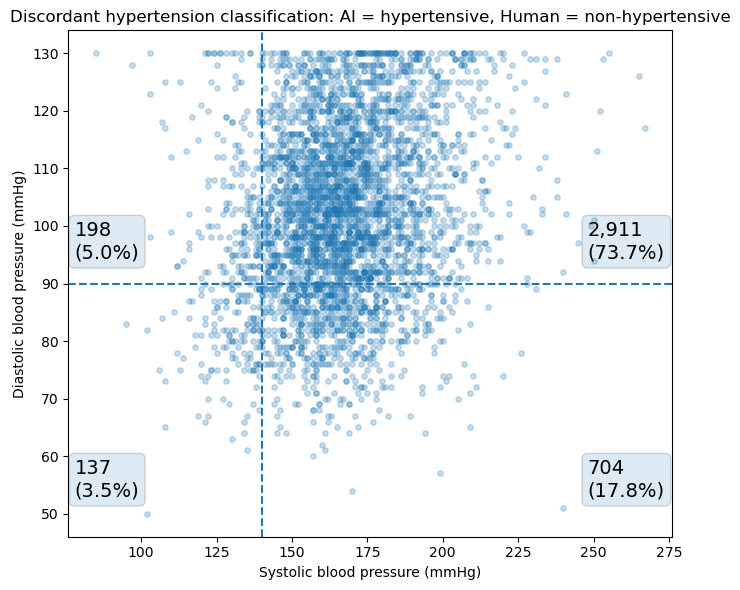}
\caption{Discordance analysis for subjects labeled hypertensive by NemoClaw
but normotensive in the expert reference cohort ($n = 3{,}950$). Scatter
plot of systolic versus diastolic blood pressure with clinical threshold
lines at 140/90 mmHg. Discordance concentrates at the boundary region
rather than reflecting random misclassification.}
\label{fig:discordance}
\end{figure}

Comparison of expert reference versus NemoClaw labels revealed
systematic discordance concentrated near clinical thresholds, not random
misclassification (\emph{Figure~\ref{fig:discordance}}). Among 3,950 subjects labeled
hypertensive by NemoClaw but normotensive in the expert reference, 2,911
(73.7\%) had systolic blood pressure $>140$ mmHg or diastolic $>90$ mmHg,
indicating clinically elevated measurements despite absent
ICD/medication labels in the expert track.

We requested a medication audit. NemoClaw generated SQL querying
antihypertensive prescriptions across angiotensin-converting enzyme inhibitors, angiotensin receptor blockers,
beta-blockers, calcium-channel blockers, diuretics, and related classes
(lisinopril through aliskiren). Only 125 of the 3,950 discordant
subjects had antihypertensive medication records, explaining much of the
disagreement: the expert cohort incorporated diagnosis and Rx history,
while the early agent track relied primarily on lab values.

We then directed the agent to exclude discordant cases without
medication evidence from the case group and rerun GWAS.
Post-reconciliation, NemoClaw cohort counts aligned closely with the
expert reference (Control 148,513 / Case 81,076). After this
step, agent-derived labels closely matched the team-defined cohort.

\begin{table}[htbp]
\centering
\caption{Label confusion: expert reference vs.\ NemoClaw (before medication reconciliation).}
\label{tab:confusion}
\begin{tabular}{llll}
\toprule
 & Expert Control & Expert Case & Expert Missing \\
\midrule
NemoClaw Control & 28,127 & 0 & 148,513 \\
NemoClaw Case & 3,950 & 24,756 & 81,076 \\
NemoClaw Missing & 15,445 & 767 & 0 \\
\bottomrule
\end{tabular}
\end{table}

\subsection{GWAS Execution and Results}

\begin{figure}[htbp]
\centering
\includegraphics[width=\linewidth]{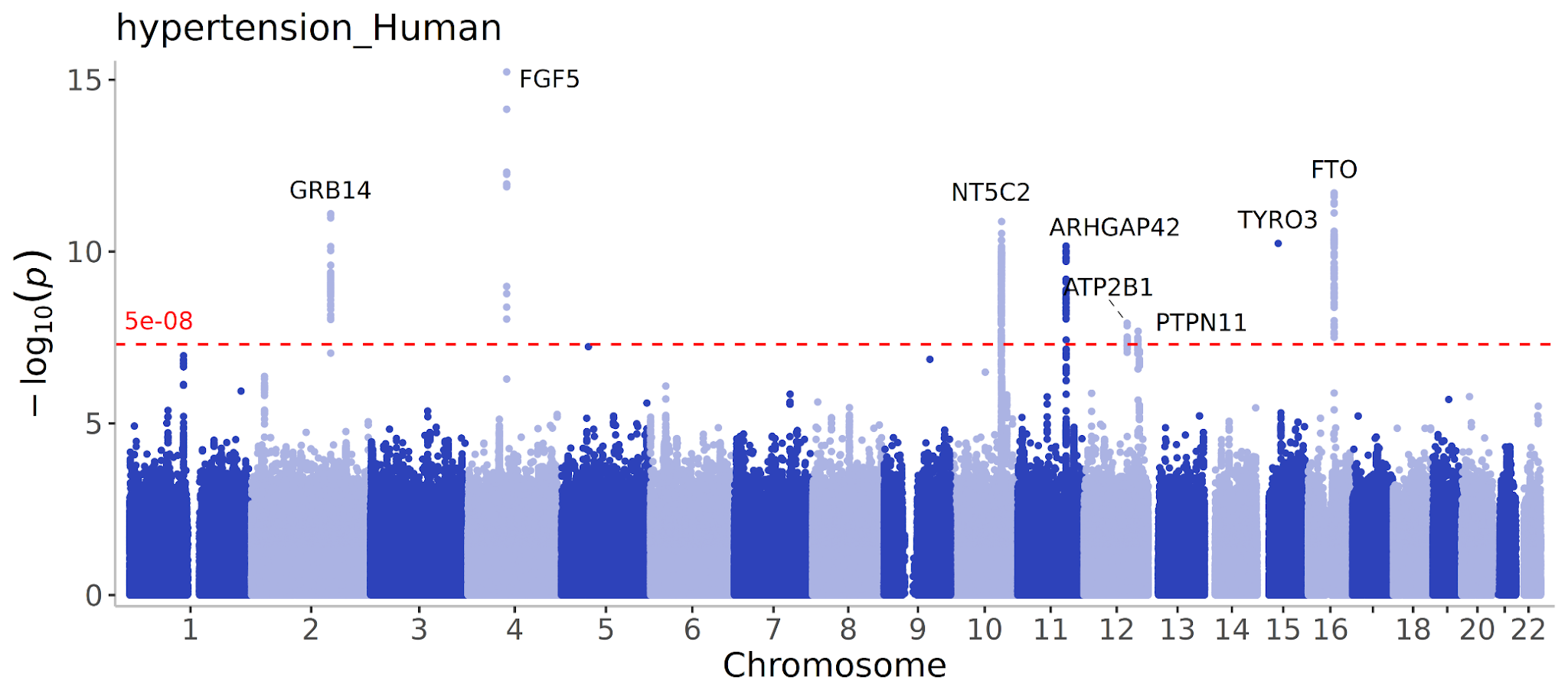}
\includegraphics[width=\linewidth]{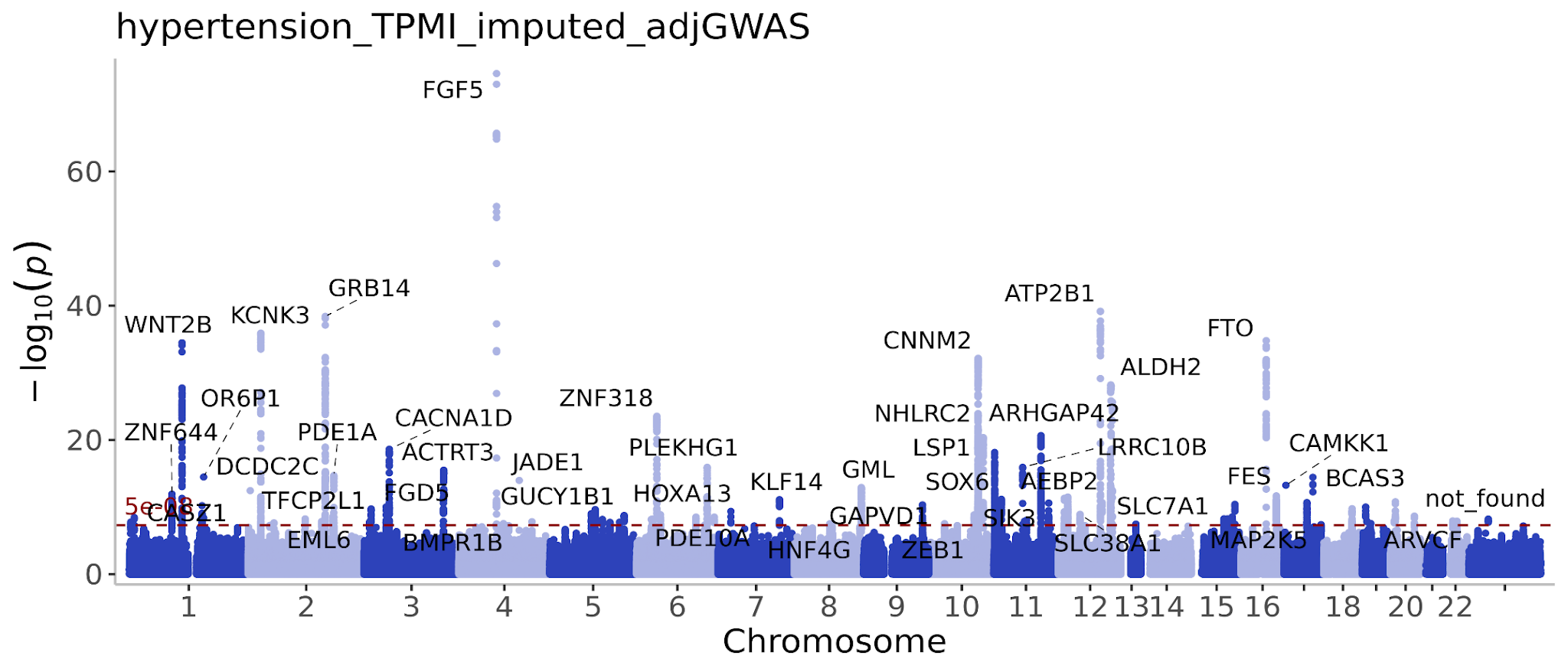}
\caption{Genome-wide association results for hypertension after phenotype
reconciliation. (Top) Manhattan plot for the expert reference cohort.
(Bottom) Manhattan plot for the NemoClaw-orchestrated GWAS. Genome-wide
significance threshold: $-\log_{10}(p) = 7.3$ ($p < 5\times10^{-8}$).
Convergent peaks at FGF5 (chr4), KCNK3/WNT2B (chr2), GRB14 (chr3),
CNNM2 (chr10), ATP2B1 (chr12), and FTO (chr16).}
\label{fig:manhattan}
\end{figure}

Both expert-curated and agent-orchestrated workflows ran PLINK2 GWAS on
the TPMI imputed library with consistent covariate adjustment.
\emph{Figure~\ref{fig:manhattan}} compares Manhattan plots for the expert reference cohort and the
NemoClaw-orchestrated GWAS. After team-directed phenotype
reconciliation, both converged on the same genome-wide significant
signals, namely FGF5, ATP2B1, CNNM2, FTO, and GRB14, demonstrating that
the agent-orchestrated analysis matched independently curated expert
results in both locus identification and signal direction. These are
replications of known hypertension loci in a Taiwanese cohort, not novel
discovery claims. Lead loci are summarized in \emph{Table~\ref{tab:loci}}.

\begin{table}[htbp]
\centering
\caption{Lead genome-wide significant loci (agent GWAS, expert-aligned phenotype).}
\label{tab:loci}
\begin{tabular}{lp{0.18\linewidth}p{0.28\linewidth}p{0.30\linewidth}}
\toprule
Chromosome & Locus/Gene & Significance (approx.\ $-\log_{10} p$) & Prior biological role \\
\midrule
4 & FGF5 & Strongest signal ($\sim$70) & Reported in European and East Asian BP GWAS \\
2 & KCNK3 / WNT2B & Genome-wide significant & Vascular tone, cardiovascular development \\
3 & GRB14 & Genome-wide significant & Metabolic/insulin signaling \\
10 & CNNM2 & Genome-wide significant & Magnesium transport, vascular reactivity \\
12 & ATP2B1 & Genome-wide significant & Vascular smooth muscle calcium transport \\
16 & FTO & Genome-wide significant & Obesity and BP regulation \\
\bottomrule
\end{tabular}
\end{table}

Significance values above the genome-wide threshold ($-\log_{10} p = 7.3$) are
read from the Manhattan plot. Exact per-single nucleotide polymorphism (SNP) effect sizes require
extraction from the underlying association result files.

Population stratification appeared well-controlled: the QQ plot returned
among the broker manifest outputs showed appropriate null distribution
at lower $p$-values, consistent with effective adjustment via PC1--PC10.
Primary outputs included per-chromosome association results, allele
frequency tables, and Manhattan and QQ visualizations, all indexed in
the broker manifest.

After team-directed phenotype reconciliation, the NemoClaw-orchestrated
GWAS matched the independently curated expert analysis in both locus
identification and signal direction, with FGF5, ATP2B1, CNNM2, FTO, and
GRB14 all replicated. This demonstrates that NAIS can produce
research-grade genomic results comparable to those of domain experts
while operating entirely within governed data boundaries and without
direct PHI access. The full pipeline, from cohort extraction and PLINK2
GWAS through Manhattan plot generation and manuscript drafting, was
planned, submitted, monitored, and documented by NemoClaw through broker
APIs without any direct PHI access at any stage.

\subsection{Expert Validation Protocol}

NemoClaw was deployed in real biomedical workflows covering literature
retrieval, workflow planning, tool selection, code generation, execution
management, result interpretation, and manuscript preparation. In the
GWAS study, the agent independently performed study planning,
statistical analysis, visualization, and manuscript drafting.
We systematically compared AI-generated workflows and outputs against
independently curated analyses on our team.

Team members remained responsible for defining scientific questions,
evaluating biological plausibility, interpreting findings, and
determining publication readiness. Validation covered SQL logic and
phenotype definitions, confusion matrices (Table~\ref{tab:confusion}) and discordance
plots (Figure~\ref{fig:discordance}), broker manifests and GLM output structure, and whether
Manhattan peaks matched domain expectations (Figure~\ref{fig:manhattan}).

Intermediate outputs were continuously validated: computational results
compared against independent analyses, code reviewed, and biological
interpretations checked against literature. Phenotype QC, not PLINK2
execution, dominated our attention during validation, illustrating where
agent value lies beyond code generation.

\subsection{Agent-Delivered Artifacts and Researcher Productivity}

NAIS accelerated the hypertension study by absorbing repetitive
orchestration that would otherwise have required specialist analyst
time: drafting and revising SQL cohort plans against the hospital
warehouse schema, generating PLINK-compatible phenotype files,
submitting and polling Kubeflow GWAS jobs through the broker, retrieving
Manhattan and QQ plots from manifests, and coordinating phenotype reruns
after discordance review. These steps proceeded through natural-language
requests with auditable broker run IDs rather than ad hoc script
maintenance. The same platform produced draft methods, results, and
presentation materials for team revision, workstreams that typically lag
primary analysis when analysts must context-switch between pipelines and
writing.

Beyond pipeline execution, NemoClaw produced research artifacts that
would typically require separate analyst effort. These included a full
manuscript draft covering the abstract, methods, results, discussion,
and tables, as well as presentation materials and slide decks. The agent
also generated workflow documentation and orchestration scripts
committed to the workspace, and maintained broker submission JSON,
phenotype TSV versions, and run\_id audit trails in the deployment
interaction logs. All artifacts required team review before any
publication use, demonstrating manuscript-oriented output under
governance rather than autonomous publication.

\subsection{Secondary Case Study: DILI Prediction}

Parallel to the GWAS work, we used NemoClaw for drug-induced liver
injury (DILI) prediction, a complementary validation that NAIS supports
multiple biomedical workflow types on the same governed agent platform.

Following literature-grounded drug list curation (33 DILI-associated and
22 low-risk compounds from clinical databases), NemoClaw developed a
progressive model architecture under team direction. Three conventional
graph-based baselines were evaluated first. Model A (Molecular Graph Attention Network),
Model B (graph neural network, GNN, with modality integration), and Model C (single-feature
logistic regression) achieved mean receiver operating characteristic area under the curve (ROC-AUC) values of 0.491--0.549, near
random performance. Incorporating literature-guided prompts in Model D
raised the mean ROC-AUC to $0.770 \pm 0.015$ (accuracy $0.700 \pm 0.015$),
demonstrating that structured prompting alone substantially improved
discriminative performance over the graph-based baselines.

\begin{figure}[htbp]
\centering
\includegraphics[width=\linewidth]{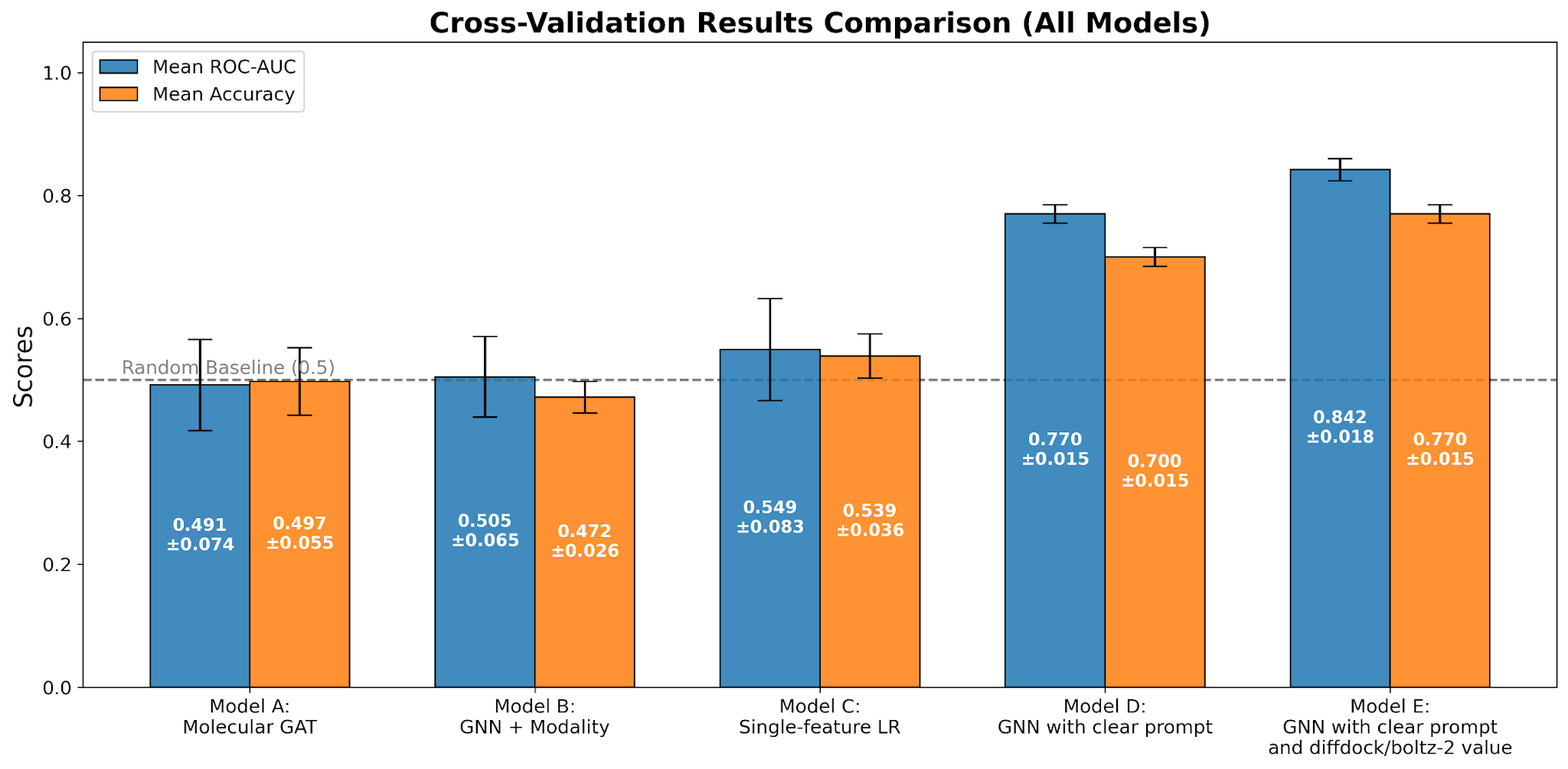}
\caption{DILI prediction model comparison. Mean AUC across multi-modal GNN
variants (Models A--E), evaluated on 390 compounds using 5-fold Murcko
scaffold cross-validation. Model E (GNN + DiffDock + Boltz-2 features
with structured prompt) achieved the highest AUC ($0.842 \pm 0.018$) and
accuracy ($0.770 \pm 0.015$).}
\label{fig:dili}
\end{figure}

We directed the research trajectory, provided compound and docking
databases, and evaluated architectural choices while NemoClaw
implemented models and cross-validation loops. A structured team prompt
specifying architectural constraints then directed NemoClaw to fuse
molecular graphs, Morgan fingerprints, physicochemical descriptors,
DiffDock binding scores, and Boltzmann-weighted binding
probabilities/IC50 vectors against PDB targets, with gating to prevent
graph-branch dominance. The resulting Model E reached AUC $0.842 \pm 0.018$
(accuracy $0.770 \pm 0.015$), demonstrating that iterative knowledge
integration on the same governed platform can substantially improve
downstream biomedical prediction performance. Figure~\ref{fig:dili} compares model
performance across Models A--E.

\section{Discussion}

The NAIS deployment surfaces three broader lessons about governed
agentic research that extend beyond the specific GWAS results. First,
agent value on protected hospital data concentrates in orchestration and
phenotype logic rather than statistical computation. Second, governance
mechanisms are not constraints on research capability but prerequisites
for institutional adoption. Third, as foundation models improve, the
binding constraint on autonomous research shifts from language-model
reasoning to AI research infrastructure. The subsections below develop
each of these in turn, followed by a frank account of the current
limitations.

\subsection{Orchestration as the Primary Agent Value}

The hypertension case study shows that agent value on protected hospital
data concentrates in orchestration and phenotype logic, not replacing
statistical genetics. PLINK2 execution is commodity. Deciding whether
hypertension means ICD codes, medications, lab thresholds, or their
combination is not. By absorbing repetitive orchestration (SQL cohort
drafting, pipeline submission and polling, plot retrieval, and rerun
coordination), NAIS freed researcher attention from script maintenance
toward study design and validation. That same reallocation of attention
is what let the 3,950-subject discordance and 73.7\% clinical-threshold
finding surface at all. A medication audit (125/3,950 on
antihypertensives) then turned the mismatch into an actionable
phenotype-design lesson rather than a silent misclassification, the kind
of finding a rushed, unaudited pipeline is more likely to bury than
catch. We observed illustrative timing gains during deployment, for
example preliminary clinical data organization in minutes versus hours
for equivalent manual workflows, but these remain observational, not
controlled benchmarks.

\subsection{Governance as Enabler}

OpenShell sandboxing, Broker v2, aggregate-only return policies, and
default network lockdown are not obstacles to research: they are
prerequisites for hospital adoption. Our deployed architecture
(Figure~\ref{fig:governed}) explicitly separates AI understanding from Broker data
governance. The agent requests analyses, Broker materializes cohorts in
ephemeral, auditable containers, and only summaries and plots return.
This enabled a 286k-person GWAS conversation that would be impossible
with an unconstrained agent holding raw EHR access.

\subsection{Autonomous Research Beyond Language Models}

Once an LLM reaches sufficient reasoning ability, the bottleneck shifts
to how the agent interacts with real research environments: reliably
executing, coordinating, and managing complex, long-running
computational workflows (data prep, statistical analysis, model
development, validation, visualization, manuscript prep) that may span
hours or days and fail due to dependencies, container incompatibility,
GPU scheduling, storage, or infrastructure interruptions. Persistent
experimental state matters more than conversational context. Future
research agents should function as orchestrators coordinating GPU
clusters, containers, workflow engines, and domain pipelines rather than
universal executors, meaning progress depends as much on AI research
infrastructure (GPU computing, containerized execution, workflow
orchestration, experiment management, reproducible environments) as on
foundation-model improvements.

The GWAS deployment illustrates this thesis in practice. Over several
weeks, NemoClaw coordinated a multi-step workflow that moved from SQL
cohort extraction to phenotype construction, Kubeflow GWAS execution,
plot generation, and manuscript drafting. Broker run IDs and deployment
logs preserved state across failures, while cohort discordance required
structured rerun orchestration rather than chat-only correction. The
DILI track showed a similar pattern: NemoClaw invoked domain pipelines
for PyTorch Geometric training, cross-validation, and multi-modal
fusion, but these capabilities depended on established computational
workflows rather than being invented from first principles by the agent.

\subsection{Limitations}

Generalizability is the primary constraint. The NAIS deployment is a
single-institution case study, and broader claims about NAIS performance
require validation across additional hospital environments and data
governance frameworks.

The GWAS results are replication evidence, not novel discovery. Lead
loci are identified qualitatively from Manhattan-plot inspection. Exact
per-SNP effect sizes and lead variant identification require extraction
from the underlying association result files, which was not performed
for this manuscript. The DILI prediction case remains illustrative.
Model E was evaluated on 390 compounds using 5-fold Murcko scaffold
cross-validation with low variance (SD 0.018), and ablation of the
DiffDock/thermodynamics branches drops AUC to 0.71, confirming the
contribution of structural features. A dataset of this size is
insufficient for clinical-grade validation (\emph{Figure~\ref{fig:dili}}).

At the system level, the proposal-review evaluator statistics represent
a modest sample of 18 production runs (Appendix A), and the variability
observed in the FP16 container execution path indicates that
governed-path coverage across diverse compute environments warrants
continued evaluation (Appendix C).

\section{Conclusion}

NAIS is a governed end-to-end architecture for
agentic research that brings together optional proposal review,
Hermes/NemoClaw execution planning, dual GPU container and
broker/Kubeflow routes, and publication-oriented evidence under
scientist-in-the-loop oversight. The primary validation is
the hypertension GWAS deployment. Through iterative agent development,
real-world pipeline debugging, and human-AI collaboration on phenotype
design, NemoClaw orchestrated a 286,422-individual GWAS under
aggregate-only data policies and produced results comparable to those of
an independently curated expert analysis, replicating established loci
including FGF5, ATP2B1, CNNM2, FTO, and GRB14. A secondary DILI workflow
reached multi-modal GNN AUC 0.842, further demonstrating NAIS capability
across biomedical workflow types on the same governed platform.

The hypertension GWAS deployment shows that NAIS can carry a real
genomics study from cohort definition through governed pipeline
execution to publication-ready results, with the research team guiding
scientific decisions throughout. When infrastructure is treated as
first-class research equipment alongside the language model itself,
governed agents can contribute meaningfully to the full research
lifecycle near protected clinical data.

\newpage
\bibliographystyle{unsrt}
\bibliography{refs}

\newpage
\appendix
\renewcommand{\thesection}{Appendix \Alph{section}}

\section{Proposal Review Pipeline and Self-Evaluating Agents}

The seven-stage proposal-readiness pipeline: (1) multimodal
document-ingestion parsing, (2) summarizer, (3) related-work retrieval
and comparison, (4) reviewer, (5) SDK recommender, (6) multi-dimensional
evaluator (novelty, technical merit, business value, ethics), (7)
template-constrained rewriter.

\begin{algorithm}[H]
\caption{Self-evaluating agent wrapper}
\begin{algorithmic}[1]
\Require $Q$, \texttt{task\_llm}, \texttt{evaluator\_llm}, optional \texttt{json\_fixer\_llm}
\State passes $\gets 0$; attempts $\gets 0$
\While{attempts $\leq$ max\_retries \textbf{and} passes $<$ min\_consecutive\_passes}
    \State $O \gets$ \texttt{task\_llm}($Q$ + any evaluator feedback)
    \If{expect\_json \textbf{and} $O$ is malformed}
        \State $O \gets$ \texttt{deterministic\_cleanup}($O$) \textbf{or} \texttt{json\_fixer\_llm}($O$)
    \EndIf
    \State $E \gets$ \texttt{evaluator\_llm}($Q$, $O$)
    \If{$E$.is\_sufficient \textbf{and} $E$.confidence $\geq$ threshold}
        \State passes $\mathrel{+}= 1$
    \Else
        \State passes $\gets 0$; feedback $\gets E$.issues $+$ $E$.suggestions
    \EndIf
    \State attempts $\mathrel{+}= 1$
\EndWhile
\State \Return verified $O$, or failure after retry exhaustion
\end{algorithmic}
\end{algorithm}

\begin{table}[htbp]
\centering
\caption{Retry behavior per stage (18 production runs).}
\label{tab:A1}
\begin{tabular}{lllll}
\toprule
Stage & Mean attempts & 1st-pass rate & Issues caught (mean) & JSON fixer used \\
\midrule
Summarizer & 4.7 & 5/18 (28\%) & 14.2 & 0/18 \\
Reviewer & 1.7 & 12/18 (67\%) & 0.8 & 1/18 \\
Proposal rewriter & 9.5 & 0/18 (0\%) & 32.3 & 13/18 (72\%) \\
\bottomrule
\end{tabular}
\end{table}

\begin{table}[htbp]
\centering
\caption{Issue categories (852 flagged deficiencies).}
\label{tab:A2}
\begin{tabular}{llr}
\toprule
Category & Count & \% \\
\midrule
Content omissions and specificity & 414 & 48.6 \\
Unmarked new content in rewriter output & 216 & 25.4 \\
Structural omissions & 85 & 10.0 \\
Vagueness or detail loss & 68 & 8.0 \\
JSON formatting errors & 37 & 4.3 \\
Hallucination or fabrication & 27 & 3.2 \\
Factual contradiction & 5 & 0.6 \\
\bottomrule
\end{tabular}
\end{table}

\begin{table}[htbp]
\centering
\caption{Score distributions across 18 proposals.}
\label{tab:A3}
\begin{tabular}{lllll}
\toprule
Dimension & Min & Max & Mean & Median \\
\midrule
Novelty & 0.0 & 100.0 & 87.3 & 85.7 \\
Technical & 0.0 & 87.5 & 72.9 & 75.0 \\
Business & 60.0 & 100.0 & 91.1 & 100.0 \\
Ethics & 28.6 & 85.7 & 58.7 & 71.4 \\
Overall & 56.6 & 90.2 & 77.5 & 78.0 \\
\bottomrule
\end{tabular}
\end{table}

\section{Execution Route Comparison}

\begin{table}[H]
\centering
\caption{NAIS execution routes.}
\label{tab:B1}
\begin{tabular}{p{0.12\linewidth}p{0.25\linewidth}p{0.25\linewidth}p{0.28\linewidth}}
\toprule
Route & Environment assumption & Agent actions & Returned artifacts \\
\midrule
GPU container workflow &
Accessible compute; containers pulled/probed/run with GPU &
Discover and launch curated GPU containers, generate code/configs, repair failures, record run state &
Logs, metrics, checkpoints, figures, tables, model outputs, result summaries, manuscript sections \\
\addlinespace
Governed Kubernetes workflow &
Protected Kubernetes/Kubeflow env with brokered action API &
Submit approved broker actions, launch/poll jobs, track params/status, retrieve controlled outputs &
Aggregate tables, plots, metrics, summaries, methods text, manuscript-ready evidence \\
\bottomrule
\end{tabular}
\end{table}

\section{GPU Container Execution Results}

\begin{table}[htbp]
\centering
\caption{Focused GPU container execution outcomes.}
\label{tab:C1}
\begin{tabular}{p{0.20\linewidth}p{0.35\linewidth}p{0.35\linewidth}}
\toprule
Workflow & Operational outcome & Preserved evidence \\
\midrule
Tabular ML workflow &
Completed matched CPU/GPU random-forest runs. All 8 baseline entries completed, and the fixed-seed repeat reproduced exactly. &
Container provenance, source/model artifacts, results, reproducibility records, logs, claim-to-evidence records \\
\addlinespace
Deep learning inference workflow &
Completed eager, compiled, FP32, and FP16 execution for the same model input. Both optimized engines built and executed. &
Immutable container provenance, serialized engine artifacts, latency records, reproducibility records, evidence mapping \\
\addlinespace
Medical imaging segmentation workflow &
Completed smoke validation and compact train-save-load-evaluate cycle on a public segmentation benchmark. &
Container provenance, training/evaluation source, logs, checkpoints, prediction artifact, manifests, results, evidence mapping \\
\bottomrule
\end{tabular}
\end{table}

The tabular ML workflow reproduced fixed-seed entries exactly. The FP32
inference re-evaluation differed by 0.62\%, and the FP16 variant differed
by 19.34\% on engine rebuild due to tactic-selection variation, recorded
as an explicit limitation.

\end{document}